\documentclass[11pt]{article}

\usepackage[final]{acl}
\usepackage{times}
\usepackage{latexsym}
\usepackage[T1]{fontenc}
\usepackage[utf8]{inputenc}
\usepackage{microtype}
\usepackage{inconsolata}
\usepackage{graphicx}

\usepackage{amsmath,hyperref}
\usepackage{amsmath,amssymb,amsfonts}
\usepackage{algorithmic}
\usepackage{textcomp}
\usepackage{xcolor}
\usepackage{amsmath}
\usepackage{bm}

\usepackage{tikz}
\usepackage{multirow}
\usepackage{float}
\usepackage{subcaption}
\usepackage{titlefoot}
\usetikzlibrary{backgrounds,decorations.pathreplacing,fit}
\DeclareMathOperator*{\argmin}{argmin}
\definecolor{ultraviolet}{HTML}{6138f5}
\newcommand{\highlight}[1]{\textcolor{ultraviolet}{\textbf{#1}}}
\newcommand{\parheader}[1]{\noindent\textbf{#1.}}

\def\BibTeX{{\rm B\kern-.05em{\sc i\kern-.025em b}\kern-.08em
    T\kern-.1667em\lower.7ex\hbox{E}\kern-.125emX}}

\title{WhisTLE:\ Deeply Supervised, Text-Only Domain Adaptation for Small Pretrained Speech Recognition Transformers}

\author{Akshat Pandey \\
    Comcast \\
    Speech AI \\
    Washington, D.C., USA \\
    \texttt{akshat\_pandey@comcast.com} \\\And
    Karun Kumar \\
    Comcast \\
    Speech AI \\
    Washington, D.C., USA \\
    \texttt{karun\_kumar@comcast} \\\And
    Raphael Tang \\
    University College London \\
    Department of Computer Science\\
    London, UK\\
    \texttt{r.tang@cs.ucl.ac.uk}
  }

\begin{document}
\maketitle

%WhisTLE: \underline{Whis}per with \underline{T}ext-Only \underline{L}exicon \underline{E}xpansion?
\begin{abstract}
Pretrained automatic speech recognition (ASR) models such as Whisper perform well but still need domain adaptation to handle unseen parlance.
In many real-world settings, collecting speech data is impractical, necessitating text-only adaptation.
We propose WhisTLE, a deeply supervised, text-only adaptation method for pretrained encoder--decoder ASR models.
WhisTLE trains a variational autoencoder (VAE) to model encoder outputs from text and fine-tunes the decoder using the learned text-to-latent encoder, optionally combined with text-to-speech (TTS) adaptation.
At inference, the original encoder is restored, incurring no extra runtime cost.
{Across four datasets and four ASR models, WhisTLE alone helps in 28 of 32 (88\%) cases, with an average relative WER drop of 22\%.
Using WhisTLE alone or in combination with other adaptation approaches helps in 116 of 144 (81\%) cases, with an average additional WER drop of 15\%.
Given all treatments that improve on baseline scores, WhisTLE helps in 45 of 55 (82\%) cases, with an average additional relative WER drop of 16\%.
WhisTLE with TTS reduces word error rate (WER) by a \textit{relative} 49\% and outperforms all non-WhisTLE baselines in 100 of 112 scenarios.}
% Across four datasets and four ASR models, WhisTLE with TTS reduces word error rate (WER) by a \textit{relative} 49.0\% and outperforms all non-WhisTLE baselines in 100 of 112 scenarios.
We also find that WhisTLE additively complements any combination of other domain adaptation approaches; we thus recommend the inclusion of WhisTLE during standard processes for adapting encoder--decoder ASR models.\footnote{Code: \url{https://github.com/akshat0123/WhisTLE}.}
    % Motivation - why do we care?
    % Production speech recognition models must be regularly fine tuned to handle
    % new or out-of-distribution vocabulary.
    % Problem statement - what is the problem?
    % The standard fine-tuning approach for the popular speech recognition model
    % Whisper~\cite{model_1} does not handle cases where we must add new
    % vocabulary to the lexicon without corresponding audio.
    % Approach - how did you go about solving the problem?
    % In this work we propose to train a text-only encoder to model Whisper encoder output
    % given text, and then we fine tune the Whisper decoder with text only using
    % our trained text-only encoder.
    % Results - what is the answer?
    % We find that our approach improves word error rate by approximately 3\% 
    % on out-of-distribution audio--text pairs without requiring audio at training
    % time.
    % When combined with other approaches to text-only adaptation, we find a
    % cumulative improvement of approximately 6\% in word error rate.
    % Conclusions - What are the implications of your answer?
    % Our work provides a generalizable approach for text-only adaptation for any
    % encoder--decoder style automatic speech recognition model.
\end{abstract}

\begin{figure}[t]
\centering
\includegraphics[width=0.76\columnwidth,page=2]{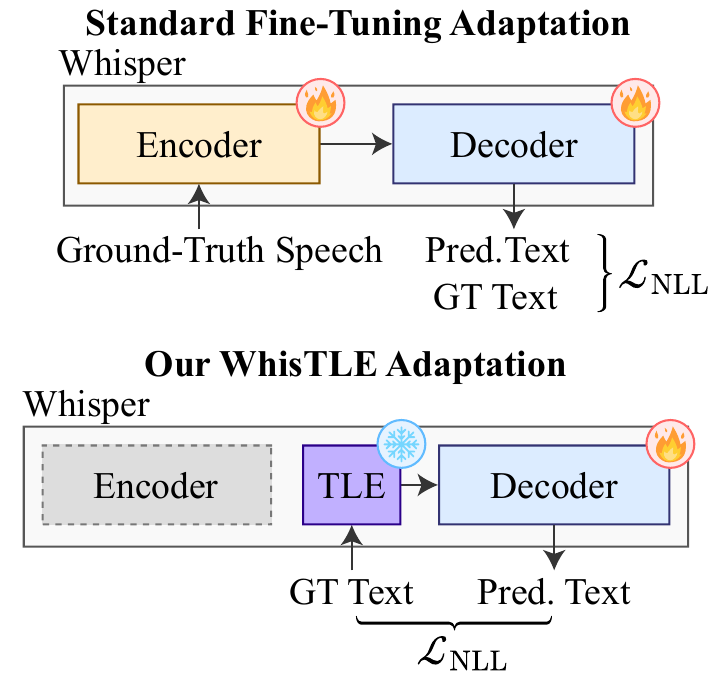}
\caption{Our WhisTLE adaptation process. 
    Speech--text pairs are treated as usual (top subfigure), while text data with no paired audio is passed through our frozen text-to-latent encoder (TLE), a trained variational auto-encoder (VAE) instead of the Whisper encoder, before being fed to the decoder (see bottom). 
    The TLE module is frozen at training time; everything else is optimized using the usual negative log-likelihood loss ($\mathcal{L}_\text{NLL}$).
}
    \label{figure:overview}
\end{figure}
\section{Introduction}
%Problem description
Although state-of-the-art automatic speech recognition (ASR) models such as Whisper~\cite{model_1} are trained on hundreds of thousands of hours of speech--text pairs, they still benefit from domain adaptation, especially if the target domain contains words unseen in the source domain.
Unfortunately, three constraints arise in practice: First, gathering speech in the target domain may be infeasible, either financially or otherwise, limiting us to text-only adaptation.
Second, the ASR system must be small (e.g., less than 1--2 billion parameters) for efficient deployment, preventing the use of speech large language models (LLMs;~\citealp{model_16}) and thus LLM text-tuning approaches~\cite{external_lm_6}.
Lastly, pretrained ASR models preclude us from applying other model architectures that afford text-only training~\cite{novel_arch_1, novel_arch_2, novel_arch_3, novel_arch_4}, as those require training from scratch.
This problem is not merely an academic artifact:\ in real-world deployment, users continuously evolve and speak new words and parlance, reducing ASR effectiveness, and collecting the required amount of speech for full training can be prohibitively expensive and time consuming.

%A standard text-only adaptation approach called shallow fusion is to train an auxiliary language model (LM) over the target domain text and linearly combine the log probabilities of the ASR model and the LM during decoding.
%However, this can still not work for unseen words since end-to-end ASR models do not generalize well to novel grapheme sequences.
A standard text-only adaptation paradigm is to train an auxiliary language model (LM) in the target domain and fuse its outputs with the ASR model.
Two popular approaches here are shallow fusion and deep fusion~\cite{external_lm_5}, with the main difference being that shallow fusion linearly combines the log probabilities of the ASR model and the LM during decoding, whereas deep fusion integrates the ASR model and the LM at the hidden feature level.
Another auxiliary model-based approach is internal language model estimation~\cite{internal_lm_3}.
Similar to shallow fusion, this approach also linearly combines the log probabilities of the ASR model and the LM during decoding, but in addition it also subtracts log probabilities produced by the ASR model's internal language model, represented by the log probabilities produced by the model after zeroing out the ASR model's acoustic representation.
However, external language modeling approaches can still fail for unseen words since end-to-end ASR models do not generalize well to novel grapheme sequences.

More promisingly, recent approaches synthesize speech using text-to-speech (TTS) models, then fine-tune the ASR model~\cite{misc_4}.
Nevertheless, TTS provides only input--output supervision and does not explicitly guide how an end-to-end model's \textit{internal state} should adapt to a target domain.
Prior work has shown deep supervision to benefit knowledge distillation~\cite{misc_2} and image classification~\cite{misc_5}, to name a few.

%Problem solution
In this work, we propose WhisTLE (adapting \underline{Whis}per with \underline{T}ext-to-\underline{L}atent \underline{E}ncodings), a deeply supervised adaptation approach for encoder--decoder ASR models, such as OpenAI's Whisper~\cite{model_1} and Canary~\cite{model_10}.
WhisTLE runs complementary to TTS adaptation, which supervises only the input speech and the output text, by also deeply supervising the ASR model's latent state.
Specifically, we train a variational autoencoder (VAE;~\citealp{model_2}) to directly model ASR encoder output using text rather than audio.
Then, we fine-tune the ASR decoder using our text-only encoder as a drop-in replacement for the ASR encoder, optionally mixing in TTS adaptation.
%\begin{figure}[t]
%\includegraphics[width=0.98\columnwidth,page=2]{sections/diagrams/00_architecture_redrawn}
%\caption{Our WhisTLE adaptation process. 
    %Speech--text pairs are treated as usual (see left), while text data with no paired audio is passed through our frozen text-to-latent encoder (TLE), a trained variational auto-encoder (VAE) instead of the Whisper encoder, before being fed to the decoder (see right). 
    %The TLE module is frozen at training time; everything else is optimized using the usual negative log-likelihood loss ($\mathcal{L}_\text{NLL}$).
%}
    %\label{figure:overview}
%\end{figure}

%Contribution
Our main contributions are as follows:\ First, to our knowledge, we propose the first deeply supervised, text-only domain adaptation method for pretrained encoder--decoder ASR transformers.
Unlike prior work on text-only training with novel and LLM architectures, our approach also applies in the pretrained, non-LLM setting.
Second, we show that WhisTLE improves ASR quality for four state-of-the-art models on out-of-distribution domains, particularly when combined with TTS adaptation, likely due to its deep latent supervision in addition to TTS's input--output supervision.
Across four out-of-domain datasets, two in-domain datasets, and four encoder--decoder models, WhisTLE with TTS yields a 49.0\% average relative WER reduction over no adaptation and outperforms all non-WhisTLE baselines in 100 out of 112 experimental scenarios.
We thus recommend the inclusion of WhisTLE in standard practices for adapting ASR models to new domains.
\section{Methodology}

% \parheaderfirst{Background}
Recently, transformer-based models~\cite{model_7} have been applied to ASR with great success~\cite{model_3, model_8, model_12}. 
Whisper~\cite{model_1} and Canary~\cite{model_10} are such state-of-the-art encoder--decoder transformers; we thus focus on them in this paper.

In the context of speech recognition, text-only domain adaptation refers to training a speech recognition model without any ground-truth audio, only text.
Current approaches can be broadly split into two categories:
those that modify and extend existing ASR models to handle new vocabulary~\cite{gen_1, gen_2, gen_3, external_lm_5, internal_lm_2, external_lm_2}, and those that detail novel ASR architectures capable of text-only training~\cite{novel_arch_1, novel_arch_2, novel_arch_3, novel_arch_4}.
We are solely concerned with the former since the latter additionally requires copious amounts of speech recognition training from scratch; we wish to work exclusively with existing ASR models, as that enables us to leverage large-scale pretrained models and greatly reduce resource consumption.
% We thus focus only on text-only adaptation rather than both text-only adaptation and speech recognition.
Literature regarding text-only domain adaptation for existing ASR models can be further split into approaches that generatively model ASR input when audio is not available~\cite{gen_3, gen_2, gen_1}, and approaches that integrate the use of an external language model at inference time~\cite{external_lm_5, internal_lm_2, external_lm_2}.

\begin{figure}[t]
    \centering
    \includegraphics[width=0.99\columnwidth,page=1]{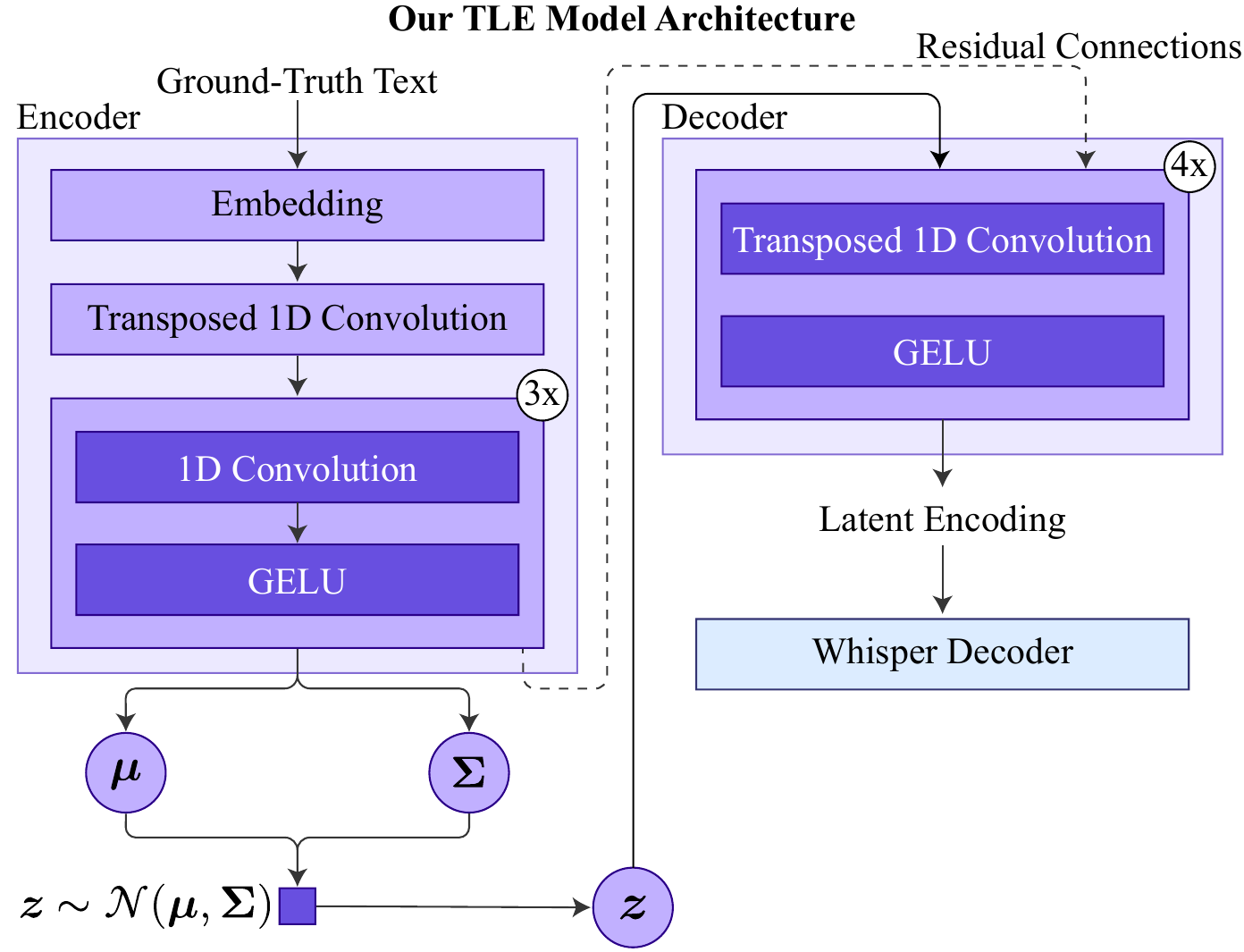}
    \caption{Model architecture of our text-to-latent encoding VAE (purple), which also follows an encoder--decoder paradigm.}
    \label{figure:architecture}
\end{figure}

\subsection{Our WhisTLE Approach}
The goals in our approach to text-only adaptation are as follows: (1) maximize our leveraging of the current state of the art in automatic speech recognition, (2) minimize training time and resource efforts, and (3) require no additional resources at inference time.
%Differentiation
Our WhisTLE approach differs from prior approaches making use of an external language model because it requires no additional resources at inference time.
It also differs from prior generative approaches in that we train a model to synthesize transformer encoder representations given text, rather than modeling audio or Mel spectrograms directly. 

Our approach rests on two hypotheses: that deep supervision of latent states helps for domain adaptation, and that modeling a transformer model's encoder output is ``simpler'' (i.e., more training efficient, as we show in Section~\ref{sec:related_work}) than modeling speech in general.
In line with the information bottleneck hypothesis~\cite{misc_1}, this is likely due to encoder representations capturing less information than speech, as they focus on what is necessary for speech recognition.

\definecolor{EncoderYellow}{HTML}{FFEAC5}
\definecolor{DecoderBlue}{HTML}{D7E8FF}
\definecolor{FrozenGrey}{HTML}{D8D8D8}
\definecolor{WhisperGrey}{HTML}{F7F7F7}
\definecolor{TLEPurple}{HTML}{B9A5FF}
\begin{figure}[h]
    \centering
    \scalebox{0.80}{
        \begin{tikzpicture}[
            box/.style={draw, minimum width=5em, minimum height=2em, align=center},
        ]
            \node[box, fill=EncoderYellow] (encoder) at (0, 0) {Encoder};
            \node[box, fill=TLEPurple, minimum width=2.5em] (tle) at (2, 0) {TLE};
            \node[box, dotted, fill=FrozenGrey] (decoder) at (4, 0) {Decoder};
            \begin{pgfonlayer}{background}
                \node[draw, fill=WhisperGrey, inner sep=0.5em, fit=(encoder) (decoder)] (wbox) {};
            \end{pgfonlayer}
            \node (speech) at (0, 1) {Speech};
            \node (text) at (2, 1) {Text};
            \node (encoding) at (0, -1) {Encoding};
            \node (pred) at (2, -1) {Pred.};
            \node (wtitle) at (4.5, 0.75) {Whisper};
            \draw[-stealth, thick] (speech) -- (encoder);
            \draw[-stealth, thick] (text) -- (tle);
            \draw[-stealth, thick] (encoder) -- (encoding);
            \draw[-stealth, thick] (tle) -- (pred);
            \node (loss) at (1, -2) {$\mathcal{L}_{VAE}$};
            \draw[decorate, decoration={brace}, thick] (2, -1.5) -- (0, -1.5);
            \node (snowflake) at (1, 0.4) {\includegraphics[scale=0.20]{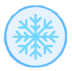}};
            \node (fire) at (2.5, 0.4) {\includegraphics[scale=0.20]{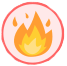}};
        \end{tikzpicture}
    }
    \caption{{Our WhisTLE adapter training process. Speech--text pairs are fed into the Whisper encoder and the TLE module, the former receiving audio and the latter receiving text. Given the ground-truth Whisper latent encoding and the predicted TLE encoding, we optimize the VAE loss.}}
    \label{fig:tle_training}
\end{figure}

Concretely, we train a variational autoencoder (VAE;~\citealp{model_2}) to precisely model Whisper~\cite{model_1} encoder outputs given text input.
After the text adaptation VAE is trained, the Whisper decoder can be fine-tuned on text without corresponding audio by using the VAE output rather than the Whisper encoder output during training. At inference time, Whisper decoding proceeds the same as it was prior to text-only adaptation, without the use of any additional runtime or memory:\ we \textit{discard} the training-only TLE module.
Figure~\ref{figure:overview} displays the pipeline for training Whisper using the text-only encoder. 
%\begin{figure}[t]
    %\centering
    %\includegraphics[width=0.98\columnwidth,page=1]{sections/diagrams/00_architecture_redrawn}
    %\caption{Model architecture of our text-to-latent encoding VAE (purple), which also follows an encoder--decoder paradigm.}
    %\label{figure:architecture}
%\end{figure}

% Formal description of methodology
Formally, let $\bm x \in \mathbb{R}^n$ be speech audio waveform in the target domain and the string $\bm y \in \Sigma^\ell$ be the tokenized transcript (of $\ell$ length) of $\bm x$, where $\Sigma$ is the vocabulary.
During full Whisper training, audio input is encoded using the Whisper
encoder $f_{\bm\theta}: \mathbb{R}^n \mapsto \mathbb{R}^{\lceil n / k \rceil\times h}$, where $k$ is the downsampling factor and $h$ is the number of hidden dimensions. 
The encodings are then fed to the decoder $g_{\bm\theta} : \mathbb{R}^{\lceil n/k \rceil \times h} \mapsto \mathbb{P}(\Sigma^{\ell})$, which outputs a differentiable probability distribution over strings of length $\ell$.
We optimize the network's parameters $\bm\theta$ end-to-end using gradient descent on backpropagating the negative log-likelihood loss, e.g.,\vspace{-1mm}
%\begin{align*}
    %\bm\theta^* &:= \argmin_{\bm\theta} \mathcal{L}_\text{NLL}, & ~\mathcal{L}_\text{NLL} := -\sum_{i=1}^\ell\log g_{\bm\theta}(f_{\bm\theta}(\bm x))[\bm y_i],
%\end{align*}
\[
\resizebox{0.65\columnwidth}{!}{$
\begin{aligned}
    \bm\theta^* &:= \argmin_{\bm\theta} \mathcal{L}_\text{NLL},\\[-1mm]
    \mathcal{L}_\text{NLL} &:= -\sum_{i=1}^\ell\log g_{\bm\theta}(f_{\bm\theta}(\bm x))[\bm y_i],
\end{aligned}
$}
\]
where $[\bm y_i]$ extracts the output probability associated with token $\bm y_i$ at the $i^\text{th}$ timestep.
% where $Y'$ are the forward shifted predicted labels given $Y$.

During text-only adaptation with WhisTLE, we instead replace $f_{\bm\theta}$ with a frozen \textit{text-based} encoder $f^\text{TLE}_{\bm\phi} : \Sigma^\ell \mapsto \mathbb{R}^{\lceil n/k \rceil \times h}$.
The Whisper encoder is set aside, and we pass only the tokens $\bm y$ to $f^\text{TLE}_{\bm\phi}$ to produce an approximation of the Whisper encoder output
$f_{\bm\theta}(\bm x)$.
We can then feed this approximation to the decoder for training and proceed with log-likelihood optimization as usual:
%\begin{align*}
    %\bm\theta_\text{TLE}^* &:= \argmin_{\bm\theta} \tilde{\mathcal{L}}_\text{NLL},\hspace{-2mm}&\tilde{\mathcal{L}}_\text{NLL} := -\sum_{i=1}^\ell\log g_{\bm\theta}(f^{\text{TLE}}_{\bm\phi}(\bm y))[\bm y_i].
%\end{align*}
\[
\resizebox{0.65\columnwidth}{!}{$
\begin{aligned}
    \bm\theta_\text{TLE}^* &:= \argmin_{\bm\theta} \tilde{\mathcal{L}}_\text{NLL},\\[-1mm]
    \tilde{\mathcal{L}}_\text{NLL} &:= -\sum_{i=1}^\ell\log g_{\bm\theta}(f^{\text{TLE}}_{\bm\phi}(\bm y))[\bm y_i].
\end{aligned}
$}
\]

\noindent Note the lack of dependence on $\bm x$.
After text-only training, we abandon $f^\text{TLE}_{\bm\phi}$ and use Whisper in the standard fashion.

Figure~\ref{figure:architecture} displays the architecture of our proposed
text-to-latent encoder (TLE).
The model is a convolutional VAE with three convolutional
encoder layers and four convolutional decoder layers, with residual connections between the respective encoder and decoder layers.
Text is embedded and upsampled using a transposed
convolutional layer prior to entering convolutional encoder layers. 

% Text only VAE pretraining
% To train this VAE, we use Whisper encoder output as labels and a
% modified version of the standard VAE loss function:
Figure~\ref{fig:tle_training} displays the training process of the TLE module.
We train our VAE on speech and text in the \textit{source} domain, $\bm x_s$ and $\bm y_s$.
We use Whisper encoder outputs as ground truth and optimize the usual VAE loss function with beta regularization:
%\begin{equation*}
    %% \mathcal{L}_\text{VAE} = \alpha \cdot\text{MSE} + \gamma \cdot \text{MSE}_\text{masked} + \lambda \cdot \text{KLDiv}.
    %\mathcal{L}_\text{VAE} := \mathbb{E} \lVert f_{\bm\theta}(\bm x) - f^\text{TLE}_{\bm\phi}(\bm y) \rVert^2_2 + \beta \text{KL}\left(\mathbb{P}_{\bm\phi}(\bm z)~\Vert~ \mathcal{N}(\bm 0, \bm I)\right), \label{equation:vae_loss}
%\end{equation*}
\[
\resizebox{0.85\columnwidth}{!}{$
\begin{aligned}
    \mathcal{L}_\text{VAE} :=&~~\mathbb{E} \lVert f_{\bm\theta}(\bm x) - f^\text{TLE}_{\bm\phi}(\bm y) \rVert^2_2 \\
    \quad& + \lambda\left[\mathbb{E} \lVert f_{\bm\theta}(\bm x_{unmasked}) - f^\text{TLE}_{\bm\phi}(\bm y) \rVert^2_2\right]\\
    \quad&+\beta \text{KL}\left(\mathbb{P}_{\bm\phi}(\bm z)~\Vert~ \mathcal{N}(\bm 0, \bm I)\right),
\end{aligned}
$}
\]
where $\mathcal{N}(\bm \mu, \bm \Sigma)$ represents the multivariate normal distribution with mean $\bm\mu$ and covariance matrix $\bm\Sigma$, the first term is the mean-squared reconstruction loss on masked audio, the second term is the mean-squared reconstruction loss on unmasked audio, scaled by $\lambda$, and the third term controls how far the posterior deviates from the isotropic unit Gaussian prior, scaling with $\beta$.
Our TLE VAE $f^\text{TLE}_{\bm\phi}$ is trained to minimize this objective, e.g., $\argmin_{\bm\phi} \mathcal{L}_\text{VAE}$.

We split the reconstruction error function of the standard variational autoencoder loss function into masked and unmasked mean-squared errors to account for the fact that the Whisper encoder does not use masking during training. 
The model must be able to learn on its own where in the original Whisper encoder output reflects input padding, and simultaneously produce accurate reconstructions of the portion of encoder output that reflects speech.

\section{Experimental Setup}
% Data
\parheader{Datasets}
We use six total datasets in our testing, two of which we consider ``in domain'' and four of which we consider ``out of domain.'' For the in-domain datasets, we use CommonVoice~\cite{data_1} ($\sim$1.1 million audio clips) and LibriSpeech~\cite{data_2} ($\sim$100k clips).
For the out-of-domain datasets, we use EMNS~\cite{data_3} ($\sim$1.1k clips), EmoV-DB~\cite{data_4} ($\sim$17k clips), the Free ST American English corpus (ST-AEDS)~\cite{data_5} ($\sim$6.8k clips), and the Open-source Multi-speaker Corpora of the English Accents in the British Isles (EABI;~\citealp{data_6}) (3.8k).

%EMNS
EMNS~\cite{data_3} is a dataset of strictly British-English acted speech focused on providing training data for storytelling, with a focus on emotional intensity.
%EmoV-DB
Similarly, EmoV-DB~\cite{data_4} also focuses on emotional speech.
%ST-AEDS
ST-AEDS~\cite{data_4} is an American English based dataset of audio transcriptions recorded on a cellphone.
%EABI
EABI~\cite{data_6} is a dataset focused on providing transcriptions from a wide breadth of English accents from the British Isles.
These choices reflect the practical setting where an end-to-end ASR model pretrained in a domain with large-scale data (e.g., CommonVoice) is adapted to a target domain with smaller resources.
In the case the target ``out of domain'' datasets diverge from the ``in domain'' datasets with regard to emotional intensity, accent, and recording environment, depending on the dataset.

% \subsection{Setup}
% Setup
We assess our WhisTLE as follows: (1) fine-tune Whisper on an in-domain dataset using the standard approach; (2) train our VAE on the same in-domain dataset; (3) train Whisper using text-only training on an out-of-domain dataset; then (4) test Whisper on both the audio and text pairs of the out-of-domain dataset. As baselines, we compare WhisTLE against TTS-generated data, shallow fusion, deep fusion, and internal language model estimation at inference time. We also experiment with combining multiple adaptation approaches, as they are not mutually exclusive with WhisTLE.

\begin{table}[t]
    \centering
    \resizebox{\columnwidth}{!}{
    \begin{tabular}{l|p{10.5cm}}
    TLE     & \textcolor{red}{loe bush} whipped him in the face and left no sting\\
    TTS     & \textcolor{red}{Lowe Bush} whipped him in the face and left no sting\\
    TLE+TTS & \textcolor{blue}{Low bush} whipped him in the face and left no sting\\\hline
    TLE     & he was defeated by \textcolor{red}{derban} in the november election\\
    TTS     & he was defeated by \textcolor{red}{derbyn} in the november election\\
    TLE+TTS & he was defeated by \textcolor{blue}{Durbin} in the november election\\\hline
    TLE     & anglican chant as a method of singing prose versions of the \textcolor{red}{sums}\\
    TTS     & Anglican chant is a method of singing prose versions of the \textcolor{red}{sums}\\
    TLE+TTS & Anglican chant is a method of singing prose versions of the \textcolor{blue}{Psalms}\\\hline
    TLE     & phillips stood undecided his ears strained to catch the \textcolor{red}{lightest} sound\\
    TTS     & philip stood undecided his ears strained to catch the \textcolor{red}{lightest} sound\\
    TLE+TTS & philip stood undecided his ears strained to catch the \textcolor{blue}{slightest} sound\\
    \end{tabular}
    }
    \caption{
    Some example differences between the top two single treatments (TLE, TTS) and their combination (TLE+TTS); improved handling of homonyms and named entity recognition is evident.
    }
    \label{table:examples}
\end{table}

\begin{table*}[t]
    \begin{minipage}{.5\textwidth}
    \centering
    \scalebox{0.66}{
    \begin{tabular}{c|c|l|c|c|c|c}\hline
        \multicolumn{2}{c|}{\multirow{2}{*}{}} & 
        \multicolumn{1}{c|}{\multirow{3}{*}{\textbf{Method}}} & 
        \multicolumn{4}{c}{\multirow{2}{*}{\textbf{Out-of-Domain Dataset}}}\\
        \multicolumn{2}{c|}{\multirow{3}{*}{}} & 
        \multicolumn{1}{c|}{\multirow{3}{*}{}} & 
        \multicolumn{4}{c}{}\\\cline{4-7}
        \multicolumn{2}{c|}{}&\multicolumn{1}{c|}{} & \textbf{EMNS} & \textbf{EmoV-DB} & \textbf{ST-AEDS} & \textbf{EABI} \\\hline
        \multirow{32}{*}{\rotatebox[origin=c]{90}{\textbf{In-Domain Dataset}}}
        &\multirow{16}{*}{\rotatebox[origin=c]{90}{\textbf{CommonVoice}}} 
        &   None          &  14.6           &  11.6           &  6.4           &  9.5           \\
        &&  \highlight{TLE}           &  9.5           &  11.8           &  3.8           &  8.5           \\
        &&  TTS           &  11.8           &  7.9           &  2.8           &  7.1           \\
        &&  SF            &  10.1           &  10.5           &  6.1           &  11.3           \\
        &&  DF            &  17.0           &  17.4           &  14.2           &  22.6           \\
        &&  ILME          &  9.6           &  10.1           &  5.1           &  4.2           \\
        &&  \highlight{TLE}+SF        &  12.4           &  8.4           &  5.3           &  6.9           \\
        &&  TTS+SF        &  8.1           &  9.0           &  3.2           &  4.7           \\
        &&  \highlight{TLE}+DF        &  16.8           &  17.2           &  11.9           &  21.2           \\
        &&  TTS+DF        &  18.3           &  12.7           &  8.2           &  17.1           \\
        &&  \highlight{TLE}+ILME      &  5.7           &  5.8           &  4.2           &  6.6           \\
        &&  TTS+ILME      &  5.2           &  6.3           &  2.9           &  2.5           \\
        &&  \highlight{TLE}+TTS       &  7.2           &  6.9           &  \textbf{2.5}  &  6.1           \\
        &&  \highlight{TLE}+TTS+SF    &  6.6           &  8.2           &  3.4           &  3.3           \\
        &&  \highlight{TLE}+TTS+DF    &  14.0           &  10.6           &  8.1           &  16.6           \\
        &&  \highlight{TLE}+TTS+ILME  &  \textbf{4.9}  &  \textbf{4.0}  &  2.8           &  \textbf{2.4}  \\[0.2ex]\cline{2-7}
        &\multirow{16}{*}{\rotatebox[origin=c]{90}{\textbf{LibriSpeech}}}
        &   None          &  11.8           &  26.4           &  7.1           &  9.4           \\
        &&  \highlight{TLE}           &  11.3           &  20.2           &  6.3           &  8.6           \\
        &&  TTS           &  10.8           &  17.5           &  \textbf{4.1}           &  7.8           \\
        &&  SF            &  84.5           &  69.8           &  52.1           &  32.9           \\
        &&  DF            &  72.2           &  58.6           &  49.2           &  40.0           \\
        &&  ILME          &  34.3           &  28.1           &  15.3           &  19.4           \\
        &&  \highlight{TLE}+SF        &  27.3           &  54.5           &  30.9           &  13.1           \\
        &&  TTS+SF        &  41.6           &  61.4           &  52.1           &  53.8           \\
        &&  \highlight{TLE}+DF        &  32.1           &  65.1           &  41.7           &  36.0           \\
        &&  TTS+DF        &  31.6           &  44.6           &  27.2           &  25.4           \\
        &&  \highlight{TLE}+ILME      &  25.6           &  28.0           &  16.1           &  11.8           \\
        &&  TTS+ILME      &  23.1           &  26.9           &  15.4           &  8.2           \\
        &&  \highlight{TLE}+TTS       &  \textbf{8.7}  &  \textbf{10.1}  &  4.2 &  \textbf{7.1}  \\
        &&  \highlight{TLE}+TTS+SF    &  26.3           &  84.7           &  72.7           &  22.0           \\
        &&  \highlight{TLE}+TTS+DF    &  24.3           &  58.6           &  27.9           &  27.7           \\
        &&  \highlight{TLE}+TTS+ILME  &  17.0           &  39.7           &  16.1           &  8.2           \\[0.2ex]\cline{1-7}
        && Standard fine-tuning & 1.0 & 1.2 & 1.0 & 2.1 \\\cline{1-7}
    \end{tabular}
    }
    \caption{Main results for Whisper-Large.}
    \label{table:whisper_large}
    \end{minipage}
    \begin{minipage}{.5\textwidth}
    \centering
    \scalebox{0.66}{
    \begin{tabular}{c|c|l|c|c|c|c}\hline
        \multicolumn{2}{c|}{\multirow{2}{*}{}} & 
        \multicolumn{1}{c|}{\multirow{3}{*}{\textbf{Method}}} & 
        \multicolumn{4}{c}{\multirow{2}{*}{\textbf{Out-of-Domain Dataset}}}\\
        \multicolumn{2}{c|}{\multirow{3}{*}{}} & 
        \multicolumn{1}{c|}{\multirow{3}{*}{}} & 
        \multicolumn{4}{c}{}\\\cline{4-7}
        \multicolumn{2}{c|}{}&\multicolumn{1}{c|}{} & \textbf{EMNS} & \textbf{EmoV-DB} & \textbf{ST-AEDS} & \textbf{EABI} \\\hline
        \multirow{32}{*}{\rotatebox[origin=c]{90}{\textbf{In-Domain Dataset}}}
        &\multirow{16}{*}{\rotatebox[origin=c]{90}{\textbf{CommonVoice}}} 
        &   None          &  16.7           &  23.1           &  13.7           &  12.5           \\
        &&  \highlight{TLE}           &  9.3           &  13.7           &  6.2           &  7.5           \\
        &&  TTS           &  5.6           &  13.5           &  4.3           &  4.8           \\
        &&  SF            &  15.2           &  23.7           &  17.3           &  17.9           \\
        &&  DF            &  19.7           &  33.4           &  22.2           &  39.9           \\
        &&  ILME          &  12.1           &  12.7           &  9.9           &  10.1           \\
        &&  \highlight{TLE}+SF        &  11.7           &  17.8           &  10.0           &  11.5           \\
        &&  TTS+SF        &  8.8           &  11.8           &  10.9           &  5.9           \\
        &&  \highlight{TLE}+DF        &  16.1           &  20.0           &  18.0           &  22.8           \\
        &&  TTS+DF        &  14.5           &  25.2           &  13.4           &  15.6           \\
        &&  \highlight{TLE}+ILME      &  7.7           &  7.4           &  8.9           &  11.5           \\
        &&  TTS+ILME      &  5.8           &  10.8           &  8.7           &  \textbf{3.4}  \\
        &&  \highlight{TLE}+TTS       &  6.1           &  8.4           &  \textbf{2.3}  &  4.0           \\
        &&  \highlight{TLE}+TTS+SF    &  6.1           &  6.4           &  6.2           &  3.9           \\
        &&  \highlight{TLE}+TTS+DF    &  14.3           &  16.8           &  12.0           &  13.3           \\
        &&  \highlight{TLE}+TTS+ILME  &  \textbf{5.5}  &  \textbf{4.6}  &  5.2           &  3.5           \\[0.2ex]\cline{2-7}
        &\multirow{16}{*}{\rotatebox[origin=c]{90}{\textbf{LibriSpeech}}}
        &   None          &  8.7           &  9.6           &  3.1           &  6.4           \\
        &&  \highlight{TLE}           &  8.3           &  7.5           &  3.0           &  5.9           \\
        &&  TTS           &  6.0           &  5.5           &  2.1           &  4.2           \\
        &&  SF            &  17.6           &  18.4           &  9.6           &  4.9           \\
        &&  DF            &  21.8           &  24.1           &  10.4           &  6.9           \\
        &&  ILME          &  17.2           &  19.0           &  9.6           &  3.4           \\
        &&  \highlight{TLE}+SF        &  16.9           &  14.0           &  11.3           &  3.2           \\
        &&  TTS+SF        &  7.4           &  9.7           &  3.6           &  2.5           \\
        &&  \highlight{TLE}+DF        &  21.4           &  22.8           &  12.1           &  6.9           \\
        &&  TTS+DF        &  19.0           &  20.4           &  10.2           &  5.3           \\
        &&  \highlight{TLE}+ILME      &  16.6           &  17.6           &  8.5           &  3.2           \\
        &&  TTS+ILME      &  6.8           &  8.6           &  3.6           &  \textbf{2.0}  \\
        &&  \highlight{TLE}+TTS       &  \textbf{5.3}  &  \textbf{4.8}  &  \textbf{2.0}  &  3.9           \\
        &&  \highlight{TLE}+TTS+SF    &  6.6           &  8.5           &  5.5           &  2.2           \\
        &&  \highlight{TLE}+TTS+DF    &  18.9           &  21.3           &  10.4           &  5.5           \\
        &&  \highlight{TLE}+TTS+ILME  &  6.3           &  7.6           &  4.2           &  2.2           \\[0.2ex]\cline{1-7}
        && Standard fine-tuning & 1.1 & 1.2 & {1.0} & {2.1} \\\cline{1-7}
    \end{tabular}
    }
    \caption{Main results for Whisper-Medium.}
    \label{table:whisper_medium}
    \end{minipage}
\end{table*}

% Standard fine tuning & toa training
\smallskip\parheader{Hyperparameters}
For in-domain fine-tuning, we train Whisper for 100K steps with a batch size of 8. 
For text-only training on out-of-domain datasets (e.g., WhisTLE and TTS adaptation), we train for 50K steps with a batch size of 8.
For every step of text-only training, we also train on the in-domain dataset for two steps using the audio and text, in order to prevent the model from catastrophically forgetting in-domain data.

% Shallow Fusion
For shallow fusion (SF; \citealp{external_lm_5}), we train a trigram language model
on the out-of-domain datasets and linearly combine it (scaled by weight $\gamma$)
with the in-domain ASR model.
For all SF experiments, we search over $\gamma$ values of 0.10, 0.25, 0.50, and 0.75, and report the best result of the set.
For deep fusion~\cite{external_lm_5} (DF), we use the same trigram language model, but train a bottleneck neural network that projects the output of the last hidden Whisper decoder layer into the same space as the language model output.
The trained output of this neural network is then used as a per-token scaling factor for the output of the language model during inference.
For experiments where CommonVoice is the ``in domain'' dataset, CommonVoice data is used to train the bottleneck network for 10k steps. 
For all LibriSpeech ``in domain'' experiments, LibriSpeech data is used to train the bottleneck network.

% Internal Language Model Estimation
For internal language model estimation (ILME; \citealp{internal_lm_3}), we again use the same trigram language model, but during decoding we add output from the language model (scaled by $\gamma$), and subtract the output from the internal language model of the ASR model (scaled by $\alpha$).
Internal language model output is calculated by submitting silence to the Whisper feature extractor, encoding it, and submitting this to the decoder to get the ``internal language model'' probabilities.
The $\gamma$ value is taken from the best performing shallow fusion model for each ``in domain'', ``out of domain'' pair, and we search over $\alpha$ values of 0.10, 0.25, 0.50, and 0.75, and report the best result of the set.

We use the FastSpeech2~\cite{model_13} and SpeechT5~\cite{model_14} models trained on CommonVoice and LibriSpeech, respectively, for the TTS adaptation experiments.
For all TTS experiments, we generate audio for out-of-domain datasets using randomly selected audio from the in-domain dataset for speaker simulation during generation.
We then fine-tune the base model on the generated audio after fine-tuning on the in-domain audio.
When combining TTS and TLE, TLE training is also performed on the same data as TTS training.
This setup aims to evaluate whether synthetic speech provides complementary supervision for improving robustness under domain shift across diverse out-of-domain evaluation conditions.
Further details for our training settings can be found in Section~\ref{sec:training_details} of the appendix.

\begin{figure*}[t]
    \centering
    \resizebox{0.95\textwidth}{!}{
    \begin{subfigure}[b]{0.49\textwidth}
        \centering
        \caption*{~~~~~~~~~~~~Whisper-Large ($\sim$1.5B parameters)}
        \begin{subfigure}[b]{0.49\textwidth}
          \includegraphics[width=\textwidth]{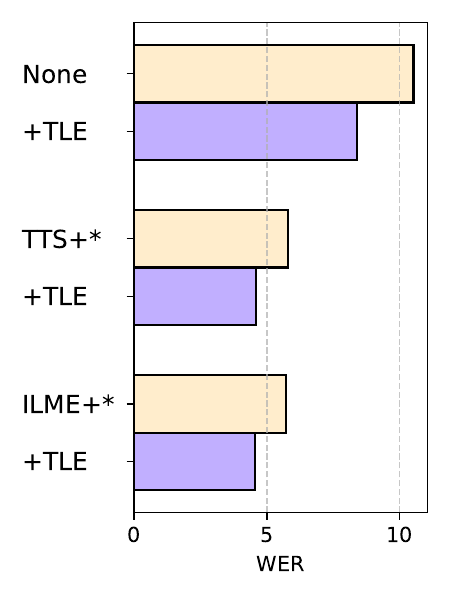}
          \captionsetup{justification=centering}
          \caption{CommonVoice\\~~~~``in domain''}
        \end{subfigure}
        \begin{subfigure}[b]{0.49\textwidth}
          \includegraphics[width=\textwidth]{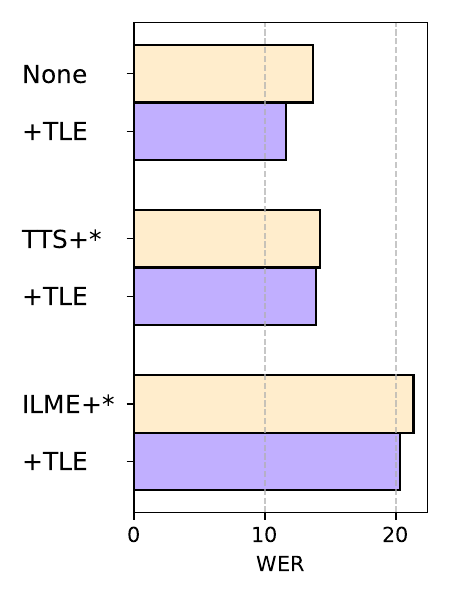}
          \captionsetup{justification=centering}
          \caption{LibriSpeech\\~~~~``in domain''}
        \end{subfigure}
    \end{subfigure}
    \setcounter{subfigure}{0}
    \begin{subfigure}[b]{0.49\textwidth}
        \centering
        \caption*{~~~~~~~~~~~~Whisper-Medium (769M parameters)}
        \begin{subfigure}[b]{0.49\textwidth}
          \includegraphics[width=\textwidth]{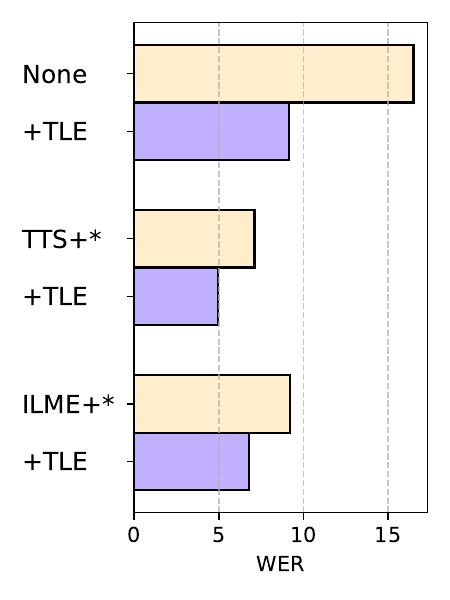}
          \captionsetup{justification=centering}
          \caption{CommonVoice\\~~~~``in domain''}
        \end{subfigure}
        \begin{subfigure}[b]{0.49\textwidth}
          \includegraphics[width=\textwidth]{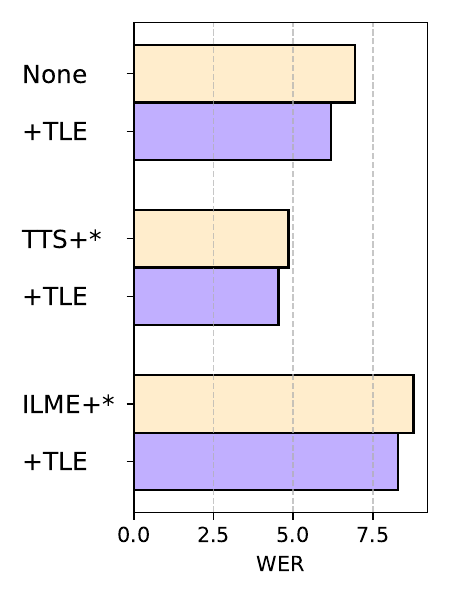}
          \captionsetup{justification=centering}
          \caption{LibriSpeech\\~~~~``in domain''}
        \end{subfigure}
    \end{subfigure}
    }
    \caption{
        The results of adding TLE as a treatment in combination with other treatments; lower WERs are better.  
        The designation ``TTS+*'' implies all treatment combinations excluding TLE, SF, and DF (i.e., TTS, TTS+ILME). The ``+TLE'' bars designate the average change in word error rate across all ``out of domain'' datasets when adding TLE to an existing treatment (i.e., TTS+TLE, TTS+ILME+TLE). Results are statistically significant at the 95\% level according to the paired bootstrap.
    }
    \label{fig:experiments}
\end{figure*}

\section{Results and Discussion}
\subsection{Main Whisper Results}
Tables~\ref{table:whisper_large} and~\ref{table:whisper_medium} display the changes in word error rates (WER) on out-of-domain datasets for our WhisTLE approach (TLE for short), training using TTS-generated data (TTS) and deep fusion (DF), inference using shallow fusion (SF) and internal language model estimation (ILME), as well as combining multiple approaches.

Overall, combining TLE and TTS attains the best average WER of 5.6 across the treatments---supporting our hypothesis that the model benefits from both input--output and latent supervision---followed by TTS at 7.2, TLE + TTS + ILME 8.4, TTS + ILME and TLE respectively at 8.8, no adaptation at 11.9, and the rest higher than 15.
Although TTS outperforms TLE, combining TLE and TTS yields an average WER improvement (1.64 points) roughly equal to the original performance gap between TTS and TLE alone (1.60), showing a strong compounding effect.
%TLE outperforms shallow fusion in 14 out of 16 occurrences, largely due to the increased occurrence of repetition hallucinations when using shallow fusion with Whisper.
TLE outperforms shallow fusion in 14 out of 16 occurrences, deep fusion in 16 out of 16 occurrences, and internal language model estimation in 12 out of 16 occurrences. Adding TLE in any combination (e.g., TLE+TTS vs.\ TTS) helps in 69 out of 80 (86\%) scenarios, resulting in an average relative WER drop of 22\%. These differences are statistically significant at the 95\% level according to the paired bootstrap.

Dataset-wise, the largest absolute gains from TLE+TTS appear on {ST-AEDS} and {EMNS}, with average drops of 3--4 WER compared to either TTS or TLE alone.
Improvements are somewhat smaller on {EABI} (2 points) and {EmoV-DB} (1.5 points) but remain consistent in direction.
Model-wise, Whisper-medium consistently benefits more from TLE+TTS, with average WERs below 6.0 on all out-of-domain sets and showing relative drops exceeding 25\% compared to no adaptation.
Whisper-large shows strong gains from TLE+TTS on CommonVoice (50\% relative drop vs.\ None), but its LibriSpeech results are noisier, partly due to SF catastrophically raising WER.
Across both models, the direction of improvement is robust, confirming that the compounding benefit of TLE and TTS is not model specific.

We also confirm that TLE additively improves any existing combination of adaptation approaches.
As shown in Figure~\ref{fig:experiments}, including TLE improves the WER on average for all in-domain dataset and method configurations, averaged across the out-of-domain datasets, with SF and DF excluded due to their quality degradation.
Adding TLE to no domain adaptation (first row) yields the largest relative improvement of 23\%, followed by any combination with TTS (15\%) and any with ILME (14\%).

In total, TLE (both alone and as part of a combination of approaches) reduces WER in 116 of 144 (81\%) cases.
Out of the 28 cases where it does not, 15 (54\%) performed worse than the baseline prior to the addition of TLE.
Of these 15 cases, all include the use of an external language modeling approach. 
Compared to TLE's reduction of WER in 81\% of cases, SF only reduces WER in 27\% of cases, DF none of the cases, and ILME 44\% of cases. 
On average, SF worsens WER by a relative 76\%, DF by 114\%, and ILME by 0\%.
Given this (along with the results of our linear-mixed effects analysis), we argue that the true underlying cause of the degradation in these 15 cases are the external language modeling methods and not TLE.  
We discuss the problems with external language modeling approaches in the context of our experiments further in Section~\ref{sec:external_lms}.

\subsection{External Language Model-Based Treatments}
\label{sec:external_lms}
Internal language model estimation (ILME) reduces average WER on out-of-domain datasets by 4.3 when the base ASR model is trained on CommonVoice. 
When the base model is trained on LibriSpeech however, ILME increases WER by an average of 8.0 points.
The LibriSpeech~\cite{data_2} corpus is a curated dataset of derived of audiobook narrations with relative stability in audio recording quality compared to CommonVoice~\cite{data_1}, which is a crowd-sourced dataset of recordings of Wikipedia text with a large variety of accents as well as variability in audio quality and formatting. 
ILME aims to alleviate domain mismatch by removing the bias induced by an ASR model's internal language model.
The difference in ILME's performance on the CommonVoice and LibriSpeech trained base models suggests that a base model trained on a dataset with more robust diversity of audio sources and quality may benefit more from an adjustment strictly limited to the ASR model's internal language model representation.

TLE- and TTS-based approaches, while less performant on LibriSpeech (TLE has an average WER drop of 1.4, and TTS 3.1), than CommonVoice (TLE has an average WER drop of 4.7, and TTS 6.3), do not suffer from the same problems as ILME in our experiments.
The TTS approach directly models audio, and TLE models the ASR model's internal audio representation, and in our experiments both are more robust to changes in audio quality in out-of-domain testing data than any external modeling based approaches we tested.

Shallow fusion and deep fusion performed the poorest in our experiments, with an increase in WER rates of 13.2 and 17.4 respectively when compared to the base trained model.
The main cause of shallow fusion's poor performance in our experiments was the dramatic increase of hallucinations produced by linearly combining the outputs of the external language model without accounting for the bias of the base model's internal language model representation (as ILME does).
Deep fusion failed largely due to the mismatch of training audio used to train the bottleneck neural network fusing the ASR and LM outputs. 
Because we are modeling a scenario where we have no access to the out-of-domain audio, the fusion network was trained on in-domain audio, which proved to be insufficient training for successfully integrating the external LM when met with out of domain audio.

\begin{table}[t]
    \setlength{\tabcolsep}{10pt}
    \centering
    \small
    \begin{tabular}{l|c|c|c}\hline
        \multicolumn{4}{c}{\textbf{Whisper-Large}}\\\hline 
        Parameter        &  Coef.   &  Std. Err.  &  $p$\\\hline
        \highlight{TLE}              &  -0.057  &  0.027     &  \textbf{0.036}\\
        ILME             &  0.072   &  0.023     &  0.002\\
        TTS              &  -0.059  &  0.019     &  0.002\\
        TLE$\times$ILME  &  -0.007  &  0.033     &  0.822\\
        TLE$\times$TTS   &  0.032   &  0.027     &  0.229\\[1ex]\hline
        \multicolumn{4}{c}{\textbf{Whisper-Medium}}\\\hline 
        Parameter     &  Coef.   &  Std. Err.  &  $p$\\\hline
        \highlight{TLE}              &  -0.034  &  0.011  &  \textbf{0.002}  \\
        ILME             &  0.004   &  0.009  &  0.636           \\
        TTS              &  -0.057  &  0.008  &  0.001           \\
        TLE$\times$ILME  &  0.022   &  0.013  &  0.096           \\
        TLE$\times$TTS   &  0.005   &  0.012  &  0.659           \\
    \end{tabular}
    \caption{Linear mixed-effects model estimation, reported using treatment coding.}
    \label{table:effectiveness_2}
\end{table}

\subsection{Linear Mixed-Effects Analysis}
%To evaluate the effectiveness of TLE across in-domain and out-of-domain datasets and to determine whether the improvement is statistically significant regardless of dataset-specific effects, we use a linear mixed-effects model as follows:
%\[
%\resizebox{0.75\columnwidth}{!}{$
%\begin{aligned}
%\text{WER}_{ijk}
%&= \beta_0 + \beta_1(\text{TLE}) \\
%&\quad + \beta_2(\text{Test Dataset}_i) \\
%&\quad + \beta_3(\text{TLE} \times \text{Test Dataset}_i) \\
%&\quad + u_j + v_k + \epsilon_{ijk},
%\end{aligned}
%$}
%\]
%where $\beta_0$ is the baseline WER, $\beta_1$ is the fixed effect for TLE, $\beta_2$ the fixed effect for the out-of-domain dataset, $\beta_3$ their interaction, $u_j$ a random intercept for the audio sample ID, $v_k$ a random intercept for the in-domain datasets, and $\epsilon_{ijk}$ the remaining residual error; all random effects are Gaussian
%This model totals 11 free parameters, estimated on 500 speech--text examples from each out-of-domain dataset.
%
%As reported in Table~\ref{table:effectiveness}, the results of the linear mixed-effects model test show that in the case of both Whisper-large and Whisper-medium, TLE reduces the reference WER (-0.8 and -6.8 points respectively) to a significant degree ($p < 0.05$). None of the interaction terms for either model are statistically significant, demonstrating the effectiveness of TLE regardless of the chosen out-of-domain dataset.
To evaluate the effectiveness of TLE across in-domain and out-of-domain datasets and to determine whether the improvement is statistically significant regardless of the effects of dataset or compounding text-only adaptation approaches, we use a linear mixed-effects model as follows:
\[
\resizebox{0.86\columnwidth}{!}{$
\begin{aligned}
\text{WER}_{ijk}
&= \beta_0 + \beta_1(\text{TLE}) + \beta_2(\text{ILME})  + \beta_3(\text{TTS}) \\
&\quad + \beta_4(\text{TLE} \times \text{ILME}) + \beta_5(\text{TLE} \times \text{TTS}) \\
&\quad + \beta_6(\text{Test Dataset}) + u_i + v_j + \epsilon_{ij},
\end{aligned}
$}
\]
where $\beta_0$ is the baseline WER, $\beta_1$ to $\beta_3$ are the fixed effects for TLE, TTS and ILME, $\beta_4$ and $\beta_5$ are the interactions between TLE and ILME and TLE and TTS, $u_i$ a random intercept for the audio sample ID, $v_j$ a random intercept for the in-domain datasets, and $\epsilon_{ij}$ the remaining residual error; all random effects are Gaussian.
Since SF and DF are prone to frequent hallucination, we omit them from these tests.
This model totals 10 free parameters, estimated on 500 speech--text examples from each out-of-domain dataset.

As reported in Table~\ref{table:effectiveness_2}, the results of the linear
mixed-effects model test show that in the case of both Whisper-large and
Whisper-medium, TLE reduces the WER (-5.7 and -3.4 points respectively) to a
significant degree ($p < 0.05$). None of the interaction terms for either model are statistically significant, demonstrating the effectiveness of TLE regardless
of the presence of other text adaptation approaches.

\begin{table}[t]
    \begin{minipage}{.5\textwidth}
    \centering
    \scalebox{0.72}{
    \begin{tabular}{c|c|l|c|c|c|c}\hline
        \multicolumn{2}{c|}{\multirow{2}{*}{}} & 
        \multicolumn{1}{c|}{\multirow{3}{*}{\textbf{Model}}} & 
        \multicolumn{4}{c}{\multirow{2}{*}{\textbf{Out-of-Domain Dataset}}}\\
        \multicolumn{2}{c|}{\multirow{3}{*}{}} & 
        \multicolumn{1}{c|}{\multirow{3}{*}{}} & 
        \multicolumn{4}{c}{}\\\cline{4-7}
        \multicolumn{2}{c|}{}&\multicolumn{1}{c|}{} & \textbf{EMNS} & \textbf{EmoV-DB} & \textbf{ST-AEDS} & \textbf{EABI} \\\hline
        \multirow{4}{*}{\rotatebox[origin=c]{90}{\textbf{In-Domain}}}
        &\multirow{2}{*}{\rotatebox[origin=c]{90}{\textbf{CV}}} 
        & Standard    & 7.8 & 18.4 &  \textbf{3.7} &  8.6 \\
        && \highlight{TLE}        & \textbf{7.4} & \textbf{16.8} &  4.5 &  \textbf{8.1} \\\cline{2-7}
        &\multirow{2}{*}{\rotatebox[origin=c]{90}{\textbf{LS}}}
        &  Standard   & 12.2 & 31.0 &  5.6 &  7.8 \\
        && \highlight{TLE}        & \textbf{11.0} & \textbf{24.5} &  \textbf{5.5} & \textbf{6.4} \\\cline{1-7}
    \end{tabular}
    }
    \caption{Results for Canary-1B.}
    \label{table:canary_large}
    \end{minipage}
\end{table}
\begin{table}[t]
    \centering
    % \vspace{3mm}
    \begin{minipage}{.5\textwidth}
    \centering
    \scalebox{0.72}{
    \begin{tabular}{c|c|l|c|c|c|c}\hline
        \multicolumn{2}{c|}{\multirow{2}{*}{}} & 
        \multicolumn{1}{c|}{\multirow{3}{*}{\textbf{Model}}} & 
        \multicolumn{4}{c}{\multirow{2}{*}{\textbf{Out-of-Domain Dataset}}}\\
        \multicolumn{2}{c|}{\multirow{3}{*}{}} & 
        \multicolumn{1}{c|}{\multirow{3}{*}{}} & 
        \multicolumn{4}{c}{}\\\cline{4-7}
        \multicolumn{2}{c|}{}&\multicolumn{1}{c|}{} & \textbf{EMNS} & \textbf{EmoV-DB} & \textbf{ST-AEDS} & \textbf{EABI} \\\hline
        \multirow{4}{*}{\rotatebox[origin=c]{90}{\textbf{In-Domain}}}
        &\multirow{2}{*}{\rotatebox[origin=c]{90}{\textbf{CV}}} 
        & Standard    & \textbf{10.4} & \textbf{10.3} & 4.9 &  9.4 \\
        && \highlight{TLE}        & 10.6 & 10.6 &  \textbf{4.7} &  \textbf{7.0} \\\cline{2-7}
        &\multirow{2}{*}{\rotatebox[origin=c]{90}{\textbf{LS}}}
        &  Standard   & 20.5 & 35.9 & 23.5 & 30.4 \\
        && \highlight{TLE}        &  \textbf{9.4} &  \textbf{9.3} &  \textbf{3.5} & \textbf{9.3} \\\cline{1-7}
    \end{tabular}
    }
    \caption{Results for Canary-180M-flash.}
    \label{table:canary_medium}
    \end{minipage}
\end{table}

\subsection{Auxiliary Canary Experiments}
To assess generalization to other encoder--decoder speech recognition models, we run experiments for Canary-1B~\cite{model_10} and Canary-180M-flash~\cite{model_11}.
Canary-1B uses a Conformer-based encoder~\cite{model_12} and a standard transformer decoder, while Canary-180M-flash is a smaller variant of Canary-1B with faster training and inference speeds. 
We choose the Canary model family for our experiments due to their comparable use of an encoder--decoder architecture, allowing for direct application of our text-only training approach.
The only adjustment made for WhisTLE is adding an additional linear layer at the end of our VAE, as Canary, unlike Whisper, also requires encoding lengths to be input into its decoder, so our VAE must also produce encoding lengths.
This is not required for Whisper~\cite{model_1}, which inherently forces encodings to account for a fixed length of thirty seconds.

% Tables~\ref{table:canary_large} and~\ref{table:canary_medium} display the
% changes in word-error rates using our WhisTLE for
% Canary-1B and Canary-180M-flash.
% When LibriSpeech~\cite{data_2} is the in-domain dataset, both the
% large and small versions of Canary perform better with text-only training.
% For the case where CommonVoice~\cite{data_1} is the in-domain dataset, text-only
% adaptation provides an improvement on 3 of 4 out-of-domain datasets for
% Canary-1b~\cite{model_10}, and an improvement on half of the out-of-domain
% datasets for Canary-180M-flash~\cite{model_11}.
Tables~\ref{table:canary_large} and~\ref{table:canary_medium} report WhisTLE results for Canary-1B and Canary-180M-flash on four out-of-domain datasets.
With LibriSpeech as the in-domain corpus, Canary-1B achieves a 16.3\% relative WER reduction and Canary-180M-flash a substantial 71\%, improving on all datasets.
These trends align with our Whisper results, where TLE consistently lowers WER.
Using CommonVoice yields smaller gains, with 4.4\% for Canary-1B and 6.0\% for Canary-180M-flash, like Whisper's more modest improvements on datasets such as EmoV-DB.
Overall, these findings suggest that text-only latent supervision generalizes beyond Whisper to other encoder--decoder ASR models.
\section{Related and Future Work}
\label{sec:related_work}
Past generative approaches include using synthetic text-to-speech (TTS) systems to generate out-of-distribution audio~\cite{gen_3} and generative adversarial networks to synthesize Mel spectrograms~\cite{gen_2}.
\citet{gen_3} rely on a closed-source paid TTS system, so model details are unavailable.
\citet{gen_2} uses 130M parameters for spectrogram generation and audio synthesis.
Our FastSpeech2~\cite{model_13} implementation has 46M parameters and SpeechT5~\cite{model_14} has 144M, both using the HiFi-GAN vocoder.
Our TLE approach uses 91M parameters for Whisper-medium and 104M for Whisper-large, which is similar in size to these TTS systems but much faster to train.
FastSpeech2 trains on batches of 48 sentences for 160K steps, while SpeechT5 jointly pretrains on text and audio with 12K-token batches before TTS fine-tuning.
Both TTS papers above also do not include the additional training required for HiFi-GAN.
In contrast, our TLE trains for only 100K steps on batches of four sentences.

A future research direction is to characterize the Pareto curve of efficiency vs. effectiveness for domain adaptation methods.
{Another is to quantify which adaptation methods may be best either alone or in combination prior to running the methods to evaluate.
Table 1 displays examples of changes in output when combining TLE and TTS treatments. 
Anecdotally, we find a combination of these approaches to lead to better handling of homonyms and named entities, but a more detailed study is needed for the exact effects of each treatment and when to use one over another.
We also feel it would be a worthwhile study to examine the changes in latent space encodings when using audio and the original model encoder versus the predicted encoding generated by the text-only adaptation module.}

%External Language Model / Combined Approaches
Using separate language models in ASR dates back to hidden Markov models and remains widely used.
In end-to-end ASR, shallow fusion~\cite{external_lm_5} combines model output scores with an external LM during inference.
\citet{internal_lm_2} approximates internal LM scores in end-to-end ASR for better external LM integration;
\citet{external_lm_2} use reranking or deep fusion with LLaMA to handle out-of-domain vocabulary.
%Architecture-based Approaches
Recent speech LLM approaches directly integrate an LLM with a speech encoder~\cite{model_16}, which allows text-only domain adaptation~\cite{external_lm_6}.
Unfortunately, the multibillion-scale nature of speech LLMs hinders our large-scale deployment to millions of end users and hence motivates the need for the present WhisTLE work.
{We also did not consider the use of large-language model-based approaches in our external language-modeling based experiments primarily because of the overhead they introduce at inference time (both in memory and runtime). 
In contrast, both TLE and TTS-based approaches introduce no additional inference-time overhead, and while n-gram models do introduce some overhead, it is minimal when compared to the prospect of loading an LLM for use as an external language model during inference.}

%Architecture-based Approaches
Finally, architecture-based approaches introduce new models to jointly learn from audio and text.
\citet{internal_lm_1} also modify Wav2vec2~\cite{model_3} and use an LLM-based decoder to reduce reliance on acoustic representations.
Hybrid Autoregressive Transducer~\cite{transducer_2} proposes a modular model with an internal LM trainable on text only.
\citet{embedding_1} add an adaptation branch embedding acoustic and linguistic features in the same space, enabling unpaired text training.
In the future, we plan to generalize our approach beyond speech recognition to any encoder--decoder task.

\section{Conclusions}
% Abstract restatement
% We introduced WhisTLE, a deeply supervised, text-only adaptation method for pretrained encoder--decoder ASR models.
% By modeling encoder outputs from text with a variational autoencoder and fine-tuning the decoder, WhisTLE provides latent-space supervision that complements the input--output supervision of TTS adaptation.
% Our experiments across four out-of-domain datasets and four ASR models show that combining WhisTLE with TTS yields a 12.3\% average relative WER reduction over TTS-only adaptation and outperforms all non-WhisTLE baselines in 27 of 32 scenarios.
% % These results demonstrate that deep supervision is an effective and efficient means of improving text-only adaptation for pretrained ASR.
% In the future, we plan to explore applying this paradigm to other sequence-to-sequence tasks and scaling WhisTLE to larger models and datasets.
We introduced WhisTLE, a deeply supervised, text-only adaptation method for pretrained encoder--decoder ASR. By modeling encoder outputs from text and fine-tuning the decoder, WhisTLE especially complements the input--output supervision of TTS adaptation. Across our experiments, WhisTLE with TTS reduced WER by a \textit{relative} 49.0\% on average and outperformed all non-WhisTLE baselines in most scenarios.
We find that WhisTLE improves WER both on its own and in combination with other adaptation approaches.
These results demonstrate the efficacy of deep latent supervision for text-only adaptation; we recommend WhisTLE as a complementary approach for leveraging unpaired text in ASR domain adaptation.
% In the future, we plan to generalize this adaptation paradigm beyond ASR.
 \section*  {Limitations}
While we have shown that the TLE approach is a significant advancement in text-only adaptation of speech recognition models, several limitations exist.
%Use of text-only adapter requires a sufficiently strong decoder
Firstly, because the method relies upon decoder training, the use of this approach requires a sufficiently large decoder to work effectively.
Models that do not have a traditional decoder, such as those that use CTC decoding~\cite{ctc} like wav2vec 2.0~\cite{wav2vec2} may not be good candidates for this approach.
%Limited out of domain datasets (north-american + british english)
Additionally, while we test our approach on multiple out-of-domain datasets, all of them are based in English and are focused on English speakers in the US and UK, as are our in-domain datasets. 
Further testing would be required to establish the effectiveness of this approach on accents that diverge further from US- and UK-based English.
%Use of method requires training text-only adapter

In direct comparison with TTS-based approaches, our method also requires training a TLE adapter specifically for a particular ASR system, whereas many trained TTS models are already available for use publicly.
{Also since TLE training is performed jointly with in-domain audio-text pairs and out-of-domain text data (as having in-domain data is a standard premise in domain adaptation), use of this approach does require access to some \textit{in-domain} audio-text data at training time.}
%Limited external language model study
Finally, due to the constraint of minimizing inference speed and cost, we limited our external language model testing to the use of an $n$-gram language model.
It is possible that using a billion-scale contemporary language model would bolster the results of using deep fusion, shallow fusion, as well as internal language model estimation in study.

%% Custom bibliography entries only

\bibliography{acl_latex}

\appendix
\section{Training Details}
\label{sec:training_details}
For all whisper-large experiments, we used a batch size of 2 with 4 gradient accumulation steps (for an equivalent batch size of 8), with a learning rate of 6.25e-8 for a total of 100k steps.
For all whisper-medium experiments, we used a batch size of 4 with 2 gradient accumulation steps (for an equivalent batch size of 8), with a learning rate of 1.25e-7 for a total of 100k steps.
For the whisper-large TLE adapter, we trained it with batch size 4 with a learning rate of 1.25e-5 for 100k steps.
For the whisper-medium TLE adapter, we trained it with batch size 4 with a learning rate of 2.5e-5 for 100k steps.
Training the TLE module for Whisper-large took approximately 12 hours, and training the TLE module for Whisper-medium took approximately 7 hours.
All training was performed on a Nvidia L40S gpu with seed 33.

For all Canary-1B experiments, we used a batch size of 4 with 2 gradient accumulation steps (for an equivalent batch size of 8) with a learning rate of 1e-6.
For all Canary-180M-flash experiments, we used a batch size of 8 with a learning rate of 1e-5.
We did not perform a full range of canary experiments to the computation requirements. 
Our primary purpose in including Canary model results is to demonstrate that TLE can work with models other than Whisper, rather than an in-depth adaptation approach comparison.

Each result in the main results tables (tables~\ref{table:whisper_large}, \ref{table:whisper_medium}, \ref{table:canary_large}, and \ref{table:canary_medium}) is the result of a single run, with seed 33.
We were unable to run multiple trials for all the experiments due to the computational requirements. 
The Whisper analysis alone involved performing 48 training runs of Whisper ([TLE, TTS, TLE+TTS] * [8 ID/OOD datasets] * [Whisper-large/medium]). This number (48) does not include deep fusion training, TTS audio generation and TLE module training.

Generating the audio for the out-of-domain data using FastSpeech2~\cite{model_13} took
approximately 6 hours, and generating audio for the out-of-domain data using
SpeechT5~\cite{model_14} took approximately 5.5 hours.

%\section{Example Appendix}
%\label{sec:appendix}

%This is an appendix.

\end{document}